\documentclass[conference]{IEEEtran}
\usepackage{cite}
\usepackage{amsmath,amssymb,amsfonts}
\usepackage{graphicx}
\usepackage{textcomp}
\usepackage{xcolor}
\usepackage{fancyhdr}
\def\BibTeX{{\rm B\kern-.05em{\sc i\kern-.025em b}\kern-.08em
    T\kern-.1667em\lower.7ex\hbox{E}\kern-.125emX}}

\begin{document}

\title{Heterogeneous Parallel Genetic Algorithm Paradigm\\
\thanks{Identify applicable funding agency here. If none, delete this.}
}

\author{\IEEEauthorblockN{Menouar Boulif}
\IEEEauthorblockA{\textit{Computer Science Department, Faculty of Sciences} \\
\textit{University M’Hamed Bougara of Boumerdes}\\
Independence Avenue, 35000, Boumerdes, Algeria. \\
boumen7@gmail.com}
}

\maketitle

\begin{abstract}
The encoding representation of the genetic algorithm can boost or hinder its performance albeit the care one can devote to operator design. Unfortunately, a representation-theory foundation that helps to find the suitable encoding for any problem has not yet become mature. Furthermore, we argue that such a best-performing encoding scheme can differ even for instances of the same problem. In this contribution, we present the basic principles of the heterogeneous parallel genetic algorithm that federates the efforts of many encoding representations in order to efficiently solve the problem in hand without prior knowledge of the best encoding.\end{abstract}


\section{Introduction}
The encoding representation has a great influence on the performance of the genetic algorithm \cite{Rothlauf2006}. In fact, encoding representations vary in their features and can have their  strength points such as minimal epistasis \cite{Beasley1993}, scalability, good blindness \cite{Boulif2006}, good redundancy \cite{Boulif2010} whereas having some limitations, such as deception \cite{Goldberg1989}, lack of scalability, bad redundancy \cite{Boulif2006}, bad blindness, etc. Actually, the encoding representation is the mean with which the GA sees its search space. Therefore, a well designed  representation eases the search for good solutions. 
Nature, gives us a lot of examples, where appropriate vision is a key issue to best performance. Indeed, a bee can easily distinguish its favourite type of flour in an abundant vegetation due to its special eyes (see Figure \ref{beeVision}). Furthermore, the bee is attracted to its reward by floral guides that are invisible to humans. 
Another example comes from snakes: in order to track a warm-blooded pray, pit vipers are equipped with infrared sensors \cite{Gracheva2010} that allows to "see" such a pray more easily.\\
\begin{figure}[h]
\centering\includegraphics[scale=0.28]{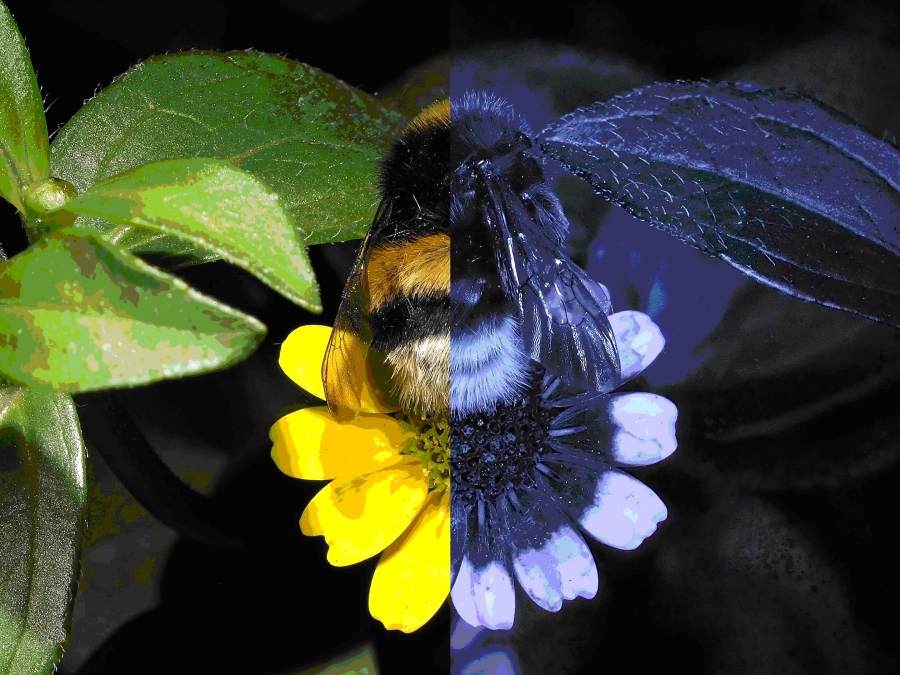} 
\caption{Human vision (left) vs bee's simulated vision that eases the detection of flours in a vegetation (right), then the discovery of nectar guided by floral guides \cite{web1}.}
\label{beeVision}
\end{figure} 
Being aware of this encoding representations differences, some researchers \cite{ronald} proposed to look for robust encoding schemes. By using nature analogy, this is to search for a type of vision that is appropriate in almost all the situations and all the needs. We think that the existence of such a kind of vision is questionable.\\
In this paper, we propose another approach for which encoding robustness is not a necessity nor a need.\\ 
The rest of this paper is organized as follows. In section 2, we present our approach. In the next section, we discuss an application of our approach to graph partitioning problem. Finaly, we present our conclusion.        

\begin{figure}[h]
\centering\includegraphics[scale=1.4]{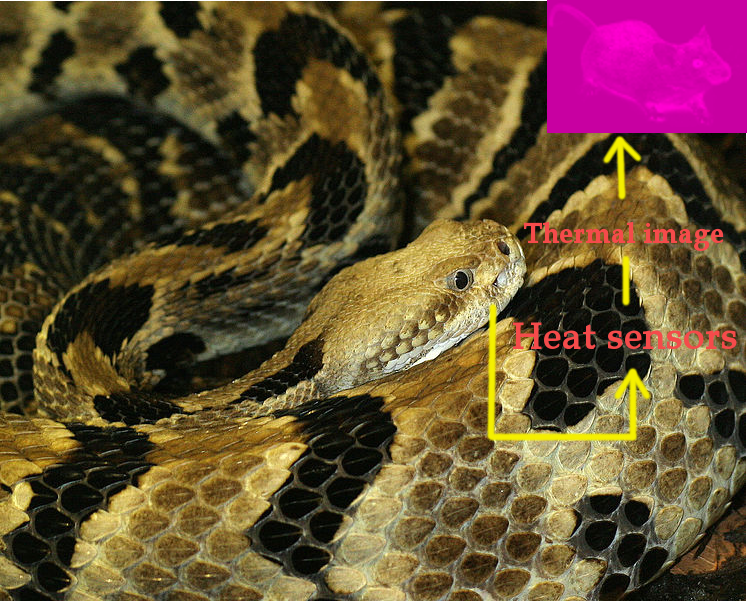} 
\caption{Pit viper \cite{web2} "sees" a thermal image of its warm-blooded pray (simulated here with Gimp\textregistered   { }tool effects).}
\label{beeVision}
\end{figure}       
 
\section{HPGA framework}
Gong et al. \cite{Gong2015} surveyed existing distributed GA and presented several related aspects such as topology, synchronization, communication, migration, etc. In contrast to the existing parallel approaches, Heterogeneous Parallel Genetic Algorithm (HPGA) combines several sub-populations but with different encoding representations. That is, HPGA uses a set of encoding schemes in order to construct a distributed genetic algorithm in which heterogeneous sub-populations evolve. This combination aims to benefit from the strength of all the representations in order to compensate their limitations. When evolving, the sub-populations share some of their best individuals according to a set of migration rules.\\   
Another aspect we think worth to explain concerns the sub-populations evolving that can be done according to a cooperative or hostile environments. For the later, the size of the best-scoring sub-populations increases to the detriment of the weak-performance islands. The score indicator of each sub-population can be calculated by using real-time performances or according to a number-of-best-solution-enhancement record. We think, this dynamic sub-population evolving allows the parallel GA to adjust itself to cope not only with the problem differences but with the instances of a same problem too.
Finally, by returning to the before mentioned natural analogy, we can notice that animal vision have a somehow kind of blindness. We think that this good blindness can play an important role in the prospecting of the search space especially when it is extremely huge. This contrasts with the wide shared belief that a good encoding scheme must be able to map the entire possible solutions. HPGA being a multiple heterogeneous sub-populations that share their information via migration, an encoding scheme that can fail to see a part of the feasible area is eligible to take part of the competition.   

\section{HPGA to solve graph partitioning}
Graph partitioning problem (GPP) consists to partition the vertex set of a graph with edge weights into subsets with some constraints (subset size for instance) so as to minimize the overall cut weight \cite{gondran}. GPP has attracted a considerable amount of research works due to its broad applicability. To solve GPP with the GA several encoding representations can be devised \cite{Chaouche2018}. 
\begin{figure}[h]
\centering\includegraphics[scale=0.4]{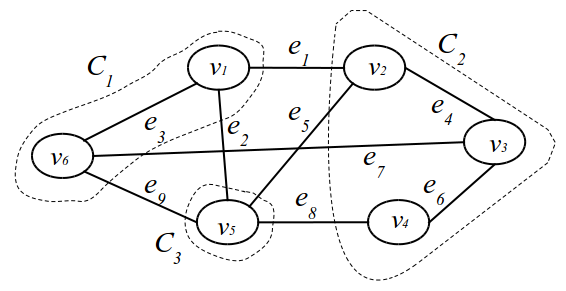} 
\caption{A three-cell solution}
\label{graph}
\end{figure}   
By using the graph G=(V,E) of Figure \ref{graph} as an example, we give in what follows a concise description of four promising encoding representations for which we refer the reader to the references for more details.  
\subsection{Fractional vertex-to-cluster encoding}
This encoding \cite{Goncalves2004} uses a fractional valued chain. Following is such a chain that represents our three-clusters partition. 

\begin{center}
\begin{tabular}[h]{|p{0.6cm} | p{0.6cm}| p{0.6cm}| p{0.6cm}| p{0.6cm}| p{0.6cm}| p{0.6cm}|}      
\hline
\centering {0.34} &\centering {0.67} &\centering {0.67} &\centering {0.67} &\centering {1.00} &\centering {0.34} & \centering {0.50}\tabularnewline
\hline
\end{tabular}
\end{center}

Each allele, except for the last one, is associated to a vertex. To get the associated partition, first we multiply the last allele by the graph order to get the number of clusters. That is, 0.5 x 6 giving us 3 clusters. Then, by calculating the product of the so obtained number of clusters times the  remaining alleles, we get the host cluster of each vertex.
\subsection{Edge based encoding}
This encoding representation \cite{Boulif2006} uses a binary vector whose alleles indicate if the edges are intra or inter-cluster. Therefore, the following chain

\begin{center}
\begin{tabular}[h]{| p{0.5 cm} | p{0.5 cm}| p{0.5 cm}| p{0.5 cm}| p{0.5 cm}| p{0.5 cm}|p{0.5 cm}| p{0.5 cm}| p{0.5 cm} |}        
\hline
\centering {1} &\centering {1} &\centering {0} &\centering {0} &\centering {1} &\centering {0}& \centering {1}& \centering {1}& \centering {1}\tabularnewline
\hline
\end{tabular}
\end{center}

defines the partition of Figure \ref{graph}. Indeed, the alleles associated to e$_3$, e$_4$ and e$_6$ are null, and hence, they are intra-cluster, whereas the remaining are inter-cluster. 
\subsection{Cut-based encoding}
This encoding representation \cite{Boulif2018} constructs a partition by its cuts, a cut being represented by a cut base (we use hereafter the nodal base defined by the first five vertices). For example, the following chain that uses at most three cuts
\begin{center}
\begin{tabular}[h]{| p{0.15 cm} | p{0.15 cm}| p{0.15 cm}| p{0.15 cm}| p{0.15 cm}|| p{0.15 cm}|p{0.15 cm}| p{0.15 cm}| p{0.15 cm}| p{0.15 cm}|| p{0.15 cm}| p{0.15 cm}| p{0.15 cm}| p{0.15 cm}| p{0.15 cm} |}        
\hline
\centering {0} &\centering {0} &\centering {0} &\centering {0} &\centering {1} &\centering {0}& \centering {1}& \centering {1}& \centering {1}& \centering {0}& \centering {0}& \centering {0}& \centering {0}& \centering {0}& \centering {0}\tabularnewline
\hline
\end{tabular}
\end{center}
defines the partition of Figure \ref{graph} by using the two cuts $\{$e$_2$,e$_5$,e$_8$,e$_9$$\}$ and $\{$e$_1$,e$_5$,e$_7$,e$_8$$\}$.  
\subsection{P-median encoding}
This encoding scheme uses the P-median approach to represent solutions \cite{Chaouche2018}. For example, the following chain
\begin{center}
\begin{tabular}[h]{| p{0.5 cm} | p{0.5 cm}| p{0.5 cm}| p{0.5 cm}| p{0.5 cm}| p{0.5 cm} |}        
\hline
\centering {0} &\centering {0} &\centering {1} &\centering {0} &\centering {1} &\centering {1}\tabularnewline
\hline
\end{tabular}
\end{center}
uses three medians, namely v$_3$, v$_5$ and v$_6$, defining three preliminary clusters. Then, an assignment procedure assigns the remaining vertices to the cluster that contains the median with which they have the most weighted direct link, yielding the partition of figure 1.  
\subsection{HPGA implemetation}

{\begin{figure*}[hbtp]
\centering\includegraphics[scale=0.6]{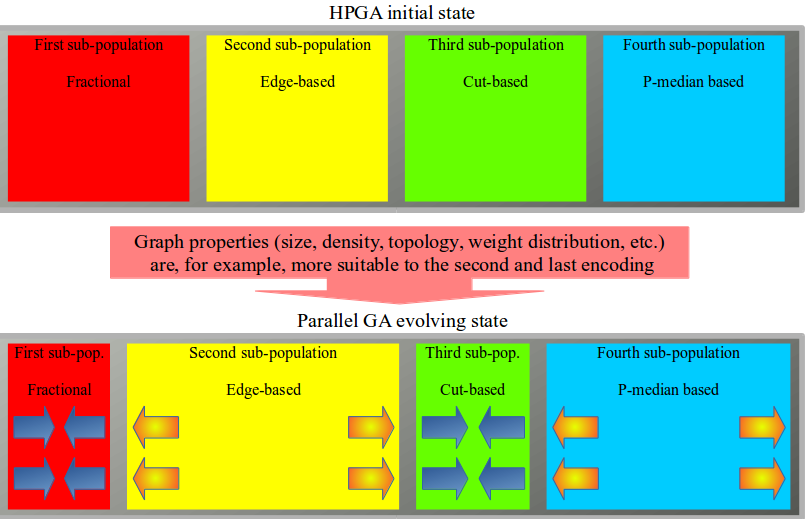} 
\caption{HPGA with dynamic sub-population sizes}
\label{hpgaFramework}
\end{figure*}}
Figure \ref{hpgaFramework} presents a possible implementation of HPGA for graph partitioning. In the initial state we choose a set of encoding schemes that embed different kind of decision variables. The subpopultaions can be equally sized or devised according to an offline processing that can guess the most promising encoding for the current instance. Afterwards, by using the performance record these subpopulations evolve dynamically to search for a good solution.\\
\begin{figure}[h]
\centering\includegraphics[scale=0.45]{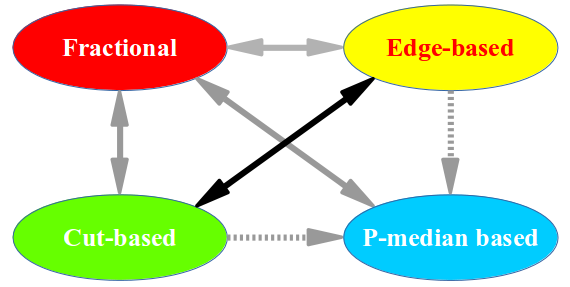} 
\caption{HPGA topology}
\label{hpgaTopology}
\end{figure}
When evolving, HPGA follows a static or dynamic topology. Figure \ref{hpgaTopology} presents a possible topology for graph partitioning that takes into account the blindness of the encoding schemes and the performance of each encoding when it is implemented alone. In fact, whereas the fractional encoding can map the entire solution space, the three others cannot. Furthermore, the P-median being the best performing representation \cite{Chaouche2018}, the flow of emigrants to it is higher than the immigrants.  
 
\section{Conclusion}
The theory of representation in evolutionary optimisation is a interesting topic that must be investigated due to the emerging of more and more challenging problems.  To the best of our knowledge, this paper propose a new paradigm that uses heterogeneous subpopulations for the parallel genetic algorithms. In addition, we present an analogy between animal blindness that helps them to survive and the non completeness of encoding schemes that can ease the prospecting of huge search spaces. Our future work focuses on applying the proposed approach to a set of intractable problems in order to verify our intuition that it will give good performances especially for large instances.    

\end{document}